\documentstyle[12pt]{article}
\newcommand{\A}{@}
\newcommand{\C}{\copyright}
\newcommand{\Bem}[1]{}
 
\date{}

\begin{document}
\large 
\quad \\
\quad \\
\quad \\
\begin{flushright}
\end{flushright}
\quad \\
\quad \\
\quad 
Mieczys{\l}aw A. K{\L}OPOTEK$^*$\footnote{$^*$ Institute of Computer Science,
Polish Academy of Sciences, PL 01-237 Warsaw, ul. Ordona 21, 
klopotek@ipipan.waw.pl} 
\quad \\
\quad \\
\quad 
 \begin{center}
\bf 
NON-DESTRUCTIVE SAMPLE GENERATION FROM CONDITIONAL BELIEF FUNCTIONS
\end{center}
\quad \\
\quad \\
\quad 
{\footnotesize
\hspace*{0.7cm}\begin{tabular}{p{16cm}}
This paper presents a new approach to generate 
samples from conditional belief functions for a restricted but non trivial 
subset of conditional belief functions.  It assumes the 
factorization (decomposition)
 of a belief function along a bayesian network structure.
It applies general conditional belief functions.
\end{tabular}
\quad \\}
\quad \\
\quad 
\noindent
{\bf 1. THE PROBLEM}\\
\quad \\
It is commonly  acknowledged that  we need to accept and handle uncertainty 
 when reasoning with real world data. 
The most profoundly studied      measure of uncertainty is the probability.
There exist methods of so-called graphoidal representation of joint 
probability distribution 
-  called Bayesian networks \cite{Pearl:88} -
allowing for expression of qualitative independence,
causality, efficient reasoning, explanation, learning from data and sample 
generation.  
However, the general feeling is that probability cannot express all types 
of 
uncertainty, including vagueness and incompleteness of knowledge. 
The Mathematical Theory of Evidence or the Dempster-Shafer Theory (DST)
\cite{
Shafer:76}
has been intensely investigated in the past as a means of expressing 
incomplete knowledge. 
\Bem{
This incompleteness means that our belief 
 into a hypothesis $A$ does not need to be fully complementary
to our belief in counter hypothesis $\overline{A}$, that is in general
$Bel(A)+Bel(\overline{A})\le 1$, whereas in the probability theory
this incompleteness could not be expressed as always $Pr(A)+Pr(
\overline{A})=1$ holds. 
}
The interesting property in this context is that DST formally fits into the 
framework of graphoidal structures \cite{Shenoy:94}
which implies possibilities of efficient reasoning by 
 local computations in large multivariate 
belief distributions given a factorization of the belief distribution  into 
low dimensional component 
conditional belief functions. This in turn qualifies DST for 
usage in expert systems dealing with uncertainty as there exist efficient 
reasoning algorithms. 
But the concept of conditional belief functions 
 is generally not usable 
for sample generation because composition of conditional belief functions
is not granted to yield joint multivariate belief distribution, as      some 
values of the belief distribution may turn out to be negative 
\cite{Klopotek:96f,Shenoy:94}. Let us illustrate the problem with Bayesian 
networks in Fig.1a) and b). Table a) below gives marginal distribution of 
$X_1$ in Fig.1a,b), table b) - conditional distributions in Fig.1a), table c) 
- conditionals in Fig.1b). 

 \begin{center}\small 
(a)%
\begin{tabular}{|   l | l |}
\hline
$X_1$ &    $m_{X_1}$ \\ \hline  
\{a\} & 0.4 \\
\{b\} & 0.4 \\
\{a,b\} & 0.2 \\
\hline \end{tabular}
%
%
 (b)%
\begin{tabular}{|   l  l | l |}
\hline
$X_i$ &    $X_{i+1}$ &    $m_{X_{i+1}|X_i}$ \\ \hline  
\{a\} $\times$  & \{a\} & 0.293333 \\ 
\{a\} $\times$  & \{b\} & -0.126667 \\
\{a\} $\times$  & \{a,b\} & -0.166667 \\
\{b\} $\times$  & \{a\} & -0.126667 \\
\{b\} $\times$  & \{b\} & 0.293333 \\
\{b\} $\times$  & \{a,b\} & -0.166667 \\
\{a,b\} $\times$  & \{a\} & 0.3 \\
\{a,b\} $\times$  & \{b\} & 0.3 \\
\{a,b\} $\times$  & \{a,b\} & 0.4 \\
\hline \end{tabular} 
 (c)%
\begin{tabular}{|   l  l | l |}
\hline
$X_1$ &    $X_{i}$ &    $m_{X_{i}|X_1}$ \\ \hline  
\{a\} $\times$  & \{a\} & 0.166667 \\
\{a\} $\times$  & \{b\} & -0.0833333 \\
\{a\} $\times$  & \{a,b\} & -0.0833333 \\
\{b\} $\times$  & \{a\} & -0.0833333 \\
\{b\} $\times$  & \{b\} & 0.166667 \\
\{b\} $\times$  & \{a,b\} & -0.0833333 \\
\{a,b\} $\times$  & \{a\} & 0.35 \\
\{a,b\} $\times$  & \{b\} & 0.35 \\
\{a,b\} $\times$  & \{a,b\} & 0.3 \\
\hline \end{tabular} \end{center}

\begin{figure}
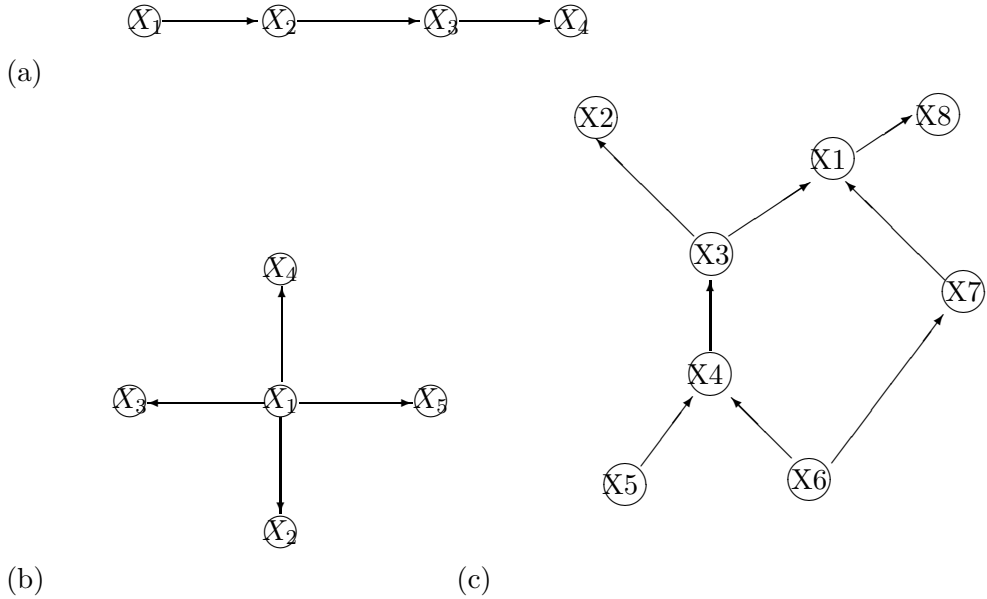

{\small 
(a) \input fig1.pic \\
(b) \input fig2.pic
(c) \input final.pic 
}
{\footnotesize
\caption{a) A chain-like bayesian network.
b) A star-like bayesian network
c) A general bayesian  network, with generated data available at
$http://www.ipipan.waw.pl/\mbox{\~{}}klopotek/ds/szampony.zip$ 
}\label{onlyone}
}
\end{figure}

In Fig.1a) $m_{X_1}
\oplus   m_{X_2|X_1}
\oplus   m_{X_3|X_2}
$ and in Fig.1b) 
 $m_{X_1}
\oplus   m_{X_2|X_1}
\oplus   m_{X_3|X_1}
\oplus   m_{X_4|X_1}
$
are proper belief functions (with non-negative values of $m$). 
But in Fig.1a) 
the function  $m=m_{X_1}
\oplus   m_{X_2|X_1}
\oplus   m_{X_3|X_2}
\oplus   m_{X_4|X_3}
$ is not a proper belief function, as visible in the table below:
\Bem{
It is easily checked that all marginals over each individual variable 
of $m=m_{X_1}
\oplus   m_{X_2|X_1}
\oplus   m_{X_3|X_2}
\oplus   m_{X_4|X_3}
$ 
are identical. 
}
 \begin{center}\small \begin{tabular}{|   l  l  l  l | l |}
\hline
X1 &    X2 &    X3 &    X4 &    m \\ \hline  
\multicolumn{4}{|c |}{.......} & ...\\
\Bem{
\{a\} $\times$  & \{a\} $\times$  & \{a\} $\times$  & \{a\} & 0.131974 \\
\{a\} $\times$  & \{a\} $\times$  & \{a\} $\times$  & \{b\} & 0.0260108 \\
\{a\} $\times$  & \{a\} $\times$  & \{a\} $\times$  & \{a,b\} & 0.0352418 \\
\{a\} $\times$  & \{a\} $\times$  & \{b\} $\times$  & \{a\} & 9.40444e-05 \\
\{a\} $\times$  & \{a\} $\times$  & \{b\} $\times$  & \{b\} & 0.0432588 \\
\{a\} $\times$  & \{a\} $\times$  & \{b\} $\times$  & \{a,b\} & 0.000353778 \\
\{a\} $\times$  & \{a\} $\times$  & \{a,b\} $\times$  & \{a\} & 0.01772 \\
\{a\} $\times$  & \{a\} $\times$  & \{a,b\} $\times$  & \{b\} & 0.01772 \\
\{a\} $\times$  & \{a\} $\times$  & \{a,b\} $\times$  & \{a,b\} & 0.0236267 \\
}
\{a\} $\times$  & \{b\} $\times$  & \{a\} $\times$  & \{a\} & 9.40444e-05 \\
\{a\} $\times$  & \{b\} $\times$  & \{a\} $\times$  & \{b\} & -2.91556e-05 \\
\{a\} $\times$  & \{b\} $\times$  & \{a\} $\times$  & \{a,b\} & -3.82222e-05 
\\ 
\multicolumn{4}{|c |}{.......} & ...\\
\hline \end{tabular} \end{center}
Also in the Fig.2b) 
the function  $m=m_{X_1}
\oplus   m_{X_2|X_1}
\oplus   m_{X_3|X_1}
\oplus   m_{X_4|X_1}
\oplus   m_{X_5|X_1}
$ is not a proper belief function as visible in the table below:
 \begin{center}\small \begin{tabular}{| lllll | l |}
\hline 
X1 &    X2 &    X3 &    X4 &    X5 &    m \\ \hline  
\multicolumn{5}{|c |}{.......} & ...\\
\Bem{
\{a\} $\times/$  & \{b\} $\times$  & \{a\} $\times$  & \{a,b\} $\times$  & 
\{a, b\} & 0.00167574 \\
\{a\} $\times$  & \{b\} $\times$  & \{b\} $\times$  & \{a\} $\times$  & \{a\} 
 & 0.00838838 \\
\{a\} $\times$  & \{b\} $\times$  & \{b\} $\times$  & \{a\} $\times$  & \{b\} 
 & 0.00287727 \\
\{a\} $\times$  & \{b\} $\times$  & \{b\} $\times$  & \{a\} $\times$  & 
\{a,b\} & 0.0022038 \\
\{a\} $\times$  & \{b\} $\times$  & \{b\} $\times$  & \{b\} $\times$  & \{a\} 
 & 0.00287727 \\
\{a\} $\times$  & \{b\} $\times$  & \{b\} $\times$  & \{b\} $\times$  & \{b\} 
 & 3.28241e-05 \\
}
\{a\} $\times$  & \{b\} $\times$  & \{b\} $\times$  & \{b\} $\times$  & 
\{a,b\} & -0.000107315 \\
\{a\} $\times$  & \{b\} $\times$  & \{b\} $\times$  & \{a,b\} $\times$  & 
\{a\} & 0.0022038 \\
\{a\} $\times$  & \{b\} $\times$  & \{b\} $\times$  & \{a,b\} $\times$  & 
\{b\} & -0.000107315 \\
\multicolumn{5}{|c |}{.......} & ...\\
\hline \end{tabular} \end{center}

Hence, in DST, sample generation from a network
and therefore the development of
learning algorithms 
 identifying  graphoidal structure from data, understanding of causality
and of mechanisms giving rise to belief distributions is hampered.  
E.g.  beside \cite{Klopotek:98}, 
 the known    sample generation algorithms 
\cite{Kaempke:88,Kreinovich:94%
,Moral:94,Moral:96,Wilson:91%
} 
do not use conditional belief
functions and  therefore (1)
conditional independence between variables cannot be pre-specified for the 
sample and (2) a single generator pass may fail to generate a single sample 
element. 
\quad \\ \quad\\

\noindent
{\bf 2. THE SOLUTION}\\
\quad \\
In our solution to the problem of sample generation from conditional 
belief functions  below  we impose 
the restriction 
 that in the bayesian network 
 no two parents of a node are directly connected.

The fundamental idea behind the approach is to replace the conditional 
belief function with a specially defined conditional probability function 
 while splitting some values of variables into subvalues. 
These subvalues take care of differences between belief function values
between subsets and supersets of elementary values of variables.  
The  proper generation of samples is run with these special conditional 
 probability functions in a very traditional way, and after completion of 
 sample generation the split values are again joined.

The main difficulties we encounter with handling      conditional belief 
functions is that 
the conditional independence in DST is radically different from probabilistic 
independence and that the conditional mass functions $m$ take negative values.

To overcome negativeness, 
we assume that the conditional belief functions are represented 
in terms of so-called $K$ functions as introduced in \cite{Klopotek:98}.
Given that $X$ is the set of all variables in the conditional belief function
and $q$ the set of conditioning variables, we have:
$$K_{|q}(A)=\sum_{B; A^{\downarrow q}\subseteq B^{\downarrow q},
 A^{\downarrow X-q} =  B^{\downarrow X-q},
} m(B)$$
 
For example, given $m$ in table (a) below, we get $K$ in table (b) below:
 \begin{center}\small 
(a)%
 \begin{tabular}{|   l  l | l |}
\hline
X1 &    X2 &    m \\ \hline  
\{a\} $\times$  & \{a\} & 0.166667 \\  
\{a\} $\times$  & \{b\} & -0.0833333 \\  
\{a\} $\times$  & \{a,b\} & -0.0833333 \\  
\{b\} $\times$  & \{a\} & -0.0833333 \\  
\{b\} $\times$  & \{b\} & 0.166667 \\  
\{b\} $\times$  & \{a,b\} & -0.0833333 \\  
\{a,b\} $\times$  & \{a\} & 0.35 \\  
\{a,b\} $\times$  & \{b\} & 0.35 \\  
\{a,b\} $\times$  & \{a,b\} & 0.3 \\  
\hline \end{tabular} 
%
 (b) %
 \begin{tabular}{|   l |l | l |}
\hline
X1 &    X2 &    $K$ \\ \hline  
\{a\} $\times$  & \{a\} & 0.516667 \\  
\{a\} $\times$  & \{b\} & 0.266667 \\  
\{a\} $\times$  & \{a,b\} & 0.216667 \\  
\{b\} $\times$  & \{a\} & 0.266667 \\  
\{b\} $\times$  & \{b\} & 0.516667 \\  
\{b\} $\times$  & \{a,b\} & 0.216667 \\  
\{a,b\} $\times$  & \{a\} & 0.35 \\  
\{a,b\} $\times$  & \{b\} & 0.35 \\  
\{a,b\} $\times$  & \{a,b\} & 0.3 \\  
\hline \end{tabular} \end{center}
$K$-function is nonnegative. For any level of conditioning variables the 
 conditioned variables form a probability distribution. 

Now we extend the set of values of every variable.
\Bem{
If the variable $X_i$ has the domain consisting of $\{a,b,c\}$, then we say 
 that 
$X_i'$ has values:
$\{a,b,c\}$, 
$\{a,b\}$,  $\{a,b\}$C$\{a,b,c\}$,  $\{a,b\}$A$\{a,b,c\}$, 
$\{b,c\}$,  $\{b,c\}$C$\{a,b,c\}$,  $\{b,c\}$A$\{a,b,c\}$, 
$\{a,c\}$,  $\{a,c\}$C$\{a,b,c\}$,  $\{a,c\}$A$\{a,b,c\}$, 
$\{a\}$, 
$\{a\}C\{a,b,c\}$, 
$\{a\}C\{a,b\}$,  $\{a\}C\{a,b\}$C$\{a,b,c\}$,  $\{a\}C\{a,b\}$A$\{a,b,c\}$, 
$\{a\}C\{a,c\}$,  $\{a\}C\{a,c\}$C$\{a,b,c\}$,  $\{a\}C\{a,c\}$A$\{a,b,c\}$, 
$\{a\}A\{a,b,c\}$, 
$\{a\}A\{a,b\}$,  $\{a\}A\{a,b\}$C$\{a,b,c\}$,  $\{a\}A\{a,b\}$A$\{a,b,c\}$, 
$\{a\}A\{a,c\}$,  $\{a\}A\{a,c\}$C$\{a,b,c\}$,  $\{a\}A\{a,c\}$A$\{a,b,c\}$.
}
If the set $S$ is a set of values of an attribute, then we define the function 
 $MY()$ as $MY(S)=S$ and $SU()$ as $SU(S)=\emptyset$. 
$S$ is a V-expression.  
For any V-expression $V$
for any proper non-empty subset $s\subset MY(V)$ we define V-expressions
$s \C V$ and $s \A V$ and define functions $MY(s \C V)=MY(s \A V)=s$,
$SU(s \C V)=SU(s \A V)=V$.
The only element of the set $\{S\}^n$ is a V(n)-expression.  
 $MY(S^n)=S$ and  $SU(S^n)=\emptyset$. 
For any V-expression $V$
for any proper non-empty subset $s\subset MY(V)$,  V(n)-expressions
are elements of the set:
$V_n$=\{$s \C V$, $s \A V$\}$^n$-\{$s \C V$\}$^n$
 and for every  $v_n\in V_n$  $MY(v_n)=s$,
$SU(v_n)=V$. Thus each V(n)-expression is a vector of $n$ V-expressions.

Let $X_j$ be a node in the belief network with $n$ successors 
and let $\pi(X_j)$ be the set of its predecessors in the network.
Let $K_{X_j |\pi(X_j)}$ be the $K$-function associated with this node.
We transform it into a conditional probability function by replacing
$X_j$ with $X_j'$ taking its   values from the set of V(n)-expressions
over the set of values of $X_j$, and  every variable $X_i\in\pi(X_j)$ 
is replaced with $X_i"$  taking its   values from the set of V-expressions
over the set of values of $X_j$. 
$P(x_j'  | x_{i1}", \dots ,x_{ik})"$ is calculated as follows:
\begin{enumerate}
\item If $SU(x_{i1}")= \dots =SU(x_{ik}")=\emptyset$ then 
for any subset of values $s$ from the domain of $X_j$ 
$\sum_{x_j';MY(x_j')=s} P(x_j'  | x_{i1}", \dots ,x_{ik}")=
K_{X_j |\pi(X_j)}(x_j'  | x_{i1}". \dots ,x_{ik}")$.
\item If $SU(x_j')\ne\emptyset$ then
$P(x_j'  | x_{i1}", \dots ,x_{ik}")=
P(SU(x_j')^n  | x_{i1}", \dots ,x_{ik}")$.
\item If $x_{il}"=MY(x_{il}")\C SU(x_{il}")$ then\\
$P(x_j'  | x_{i1}", \dots, x_{il}", \dots ,x_{ik}")=
 P(x_j'  | x_{i1}", \dots, SU(x_{il}"), \dots ,x_{ik}")$
\item 
if $x_{il}"=MY(x_{il}")\A SU(x_{il}")$ 
let $x_{il}^*$ denote either $x_{il}"$ or $SU(x_{il}")$ and otherwise
let $x_{il}^*$ denote only $x_{il}"$. 
if $x_{il}"=MY(x_{il}")\A SU(x_{il}")$ 
let $x_{il}^+$ denote either  $MY(x_{il}")$ and otherwise
let $x_{il}^+$ denote only $x_{il}"$. 
Then \\
$P(x_j'  | x_{i1}^+, \dots, x_{ik}^+)=
average_{x_{i1}^*,  \dots ,x_{ik}^*}      (
 P(x_j'  | x_{i1}^*,  \dots ,x_{ik}^*) )$ 
\end{enumerate} 
Obviously $P(x_j'  | x_{i1}", \dots ,x_{ik}")$ 
has to be non-negative everywhere.\\
If $X_j$ is a parent of another node in the network on the $h-th$ outgoing 
edge, then the respective $x_j"$ acts as the $h-th$ element of the vector 
$x_j'$. 

With such a transformed probability distribution we generate the sample and 
then replace all the V- and V(n) expressions $V$ with $MY(V)$.

If  $X2$ has  a single successor and $K_{X2|X1}$ is of the form 
 \begin{center}\small \begin{tabular}{|   l |l | l |}
\hline
X1 &    X2 &    $K$ \\ \hline  
\{a\} $\times$  & \{a\} & 0.516667 \\  
\{a\} $\times$  & \{b\} & 0.266667 \\  
\{a\} $\times$  & \{a,b\} & 0.216667 \\  
\{b\} $\times$  & \{a\} & 0.266667 \\  
\{b\} $\times$  & \{b\} & 0.516667 \\  
\end{tabular}
 \begin{tabular}{|   l |l | l |}
\hline
X1 &    X2 &    $K$ \\ \hline  
\{b\} $\times$  & \{a,b\} & 0.216667 \\  
\{a,b\} $\times$  & \{a\} & 0.35 \\  
\{a,b\} $\times$  & \{b\} & 0.35 \\  
\{a,b\} $\times$  & \{a,b\} & 0.3 \\  
\hline \end{tabular} \end{center}
then the above rules lead to $P(X2' | X1')$ of the form 
 \begin{center}\small \begin{tabular}{|   l |l | l |}
\hline
X1" &    X2' &    $P$ \\ \hline  
\{a\} $\times$  & \{a\} & 0.3               \\  
\{a\} $\times$  & \{a\}\A\{a,b\}  &0.216667 \\  
\{a\} $\times$  & \{b\} & 0.05             \\  
\{a\} $\times$  & \{b\}\A\{a,b\} & 0.216667 \\  
\{a\} $\times$  & \{a,b\} & 0.216667 \\  
\{b\} $\times$  & \{a\} & 0.05              \\  
\{b\} $\times$  & \{a\}\A\{a,b\} & 0.216667 \\  
\{b\} $\times$  & \{b\} & 0.3               \\  
\{b\} $\times$  & \{b\}\A\{a,b\} & 0.216667 \\  
\{b\} $\times$  & \{a,b\} & 0.216667 \\  
\{a,b\} $\times$  & \{a\} & 0.05     \\  
\{a,b\} $\times$  & \{a\}\A\{a,b\} & 0.3 \\  
\{a,b\} $\times$  & \{b\} & 0.05      \\  
\{a,b\} $\times$  & \{b\}\A\{a,b\} & 0.3 \\  
\{a,b\} $\times$  & \{a,b\} & 0.3 \\  
\end{tabular}
\begin{tabular}{|   l |l | l |}
\hline
X1" &    X2' &    $P$ \\ \hline  
\{a\}\C\{a,b\} $\times$  & \{a\} & 0.05 \\  
... & ... & ...\\
\{b\}\C\{a,b\} $\times$  & \{a\} & 0.05 \\  
... & ... & ...\\
\{b\}\C\{a,b\} $\times$  & \{a,b\} & 0.3 \\  
%
\{a\}\A\{a,b\} $\times$  & \{a\} & 0.55 \\  
\{a\}\A\{a,b\} $\times$  & \{a\}\A\{a,b\} & 0.133333 \\  
\{a\}\A\{a,b\} $\times$  & \{b\} & 0.05  \\  
\{a\}\A\{a,b\} $\times$  & \{b\}\A\{a,b\} &  0.133333  \\  
\{a\}\A\{a,b\} $\times$  & \{a,b\} &  0.133333  \\  
\{b\}\A\{a,b\} $\times$  & \{a\} & 0.05 \\  
\{b\}\A\{a,b\} $\times$  & \{a\}\A\{a,b\} &  0.133333  \\  
\{b\}\A\{a,b\} $\times$  & \{b\} & 0.55  \\  
\{b\}\A\{a,b\} $\times$  & \{b\}\A\{a,b\} &  0.133333  \\  
\{b\}\A\{a,b\} $\times$  & \{a,b\} &  0.133333  \\  
\hline \end{tabular} \end{center}

To verify the above sample generation 
 algorithm, a program has been implemented allowing to generate 
the sample from conditional belief functions and to test DST  conditional 
 independence properties of the sample. 
The independence test is based on a previously elaborated layered 
 independence test \cite{Klopotek:96f}. 
The PC algorithm of Spirtes/Glymour/Scheines
\cite{Spirtes:93}
 has been successfully tested   for multivariate  belief distributions for 
 samples generated by our approach. Fig.1c) represents one of the networks
recovered.

\newcommand{\LitStelle}[2]{\bibitem{#1}}  
\renewcommand{\A}[2]{\uppercase{#2 } #1} 


\newcommand{\eng}[1]{}    
\newcommand{\pol}[1]{}    
\newcommand{\deu}[1]{}    

\renewcommand{\eng}[1]{#1}    

\newcommand{\IN}{\pol{[w:]}\deu{[in:]}\eng{[in:]}} 

\newcommand{\ReadingsIn}{G. Shafer, J. Pearl eds: 
{\it Readings in Uncertain 
Reasoning}, (ISBN 1-55860-125-2, 
Morgan Kaufmann Publishers Inc., San Mateo, California, 1990)}

\end{document}